\documentclass[letterpaper, 10 pt, conference]{ieeeconf/ieeeconf}  

\IEEEoverridecommandlockouts                              

\usepackage{graphicx}
\usepackage{comment}
\usepackage{amsmath,amssymb} 
\usepackage{color}

\usepackage{booktabs}

\usepackage{cite}

\usepackage[para]{threeparttable}

\DeclareMathOperator*{\argmin}{arg\,min}

\title{\LARGE \bf
Robust Planning for Autonomous Driving via Mixed Adversarial Diffusion Predictions
}

\author{Albert Zhao$^{1}$ and Stefano Soatto$^{1}$
\thanks{© 2025 IEEE.  Personal use of this material is permitted.  Permission from IEEE must be obtained for all other uses, in any current or future media, including reprinting/republishing this material for advertising or promotional purposes, creating new collective works, for resale or redistribution to servers or lists, or reuse of any copyrighted component of this work in other works.}%
\thanks{This work has been supported by grant ONR N00014-22-1-2252.}%
\thanks{$^{1}$The authors are with the Samueli School of Engineering, Computer Science Department, University of California,
        Los Angeles, CA 90095, USA.
        {\tt\small azzhao@cs.ucla.edu, soatto@cs.ucla.edu}. 
        Corresponding author: Albert Zhao (\tt\small azzhao@cs.ucla.edu)}%
}

\begin{document}

\maketitle
\thispagestyle{empty}
\pagestyle{empty}

\begin{abstract}

We describe a robust planning method for autonomous driving that mixes normal and adversarial agent predictions output by a diffusion model trained for motion prediction. We first train a diffusion model to learn an unbiased distribution of normal agent behaviors. We then generate a distribution of adversarial predictions by biasing the diffusion model at test time to generate predictions that are likely to collide with a candidate plan. We score plans using expected cost with respect to a mixture distribution of normal and adversarial predictions, leading to a planner that is robust against adversarial behaviors but not overly conservative when agents behave normally. Unlike current approaches, we do not use risk measures that over-weight adversarial behaviors while placing little to no weight on low-cost normal behaviors or use hard safety constraints that may not be appropriate for all driving scenarios. We show the effectiveness of our method on single-agent and multi-agent jaywalking scenarios as well as a red light violation scenario.

\end{abstract}


\section{Introduction}

\label{sec:intro}

Predicting agent behaviors~\cite{gao2020vectornet,salzmann2020trajectron++,shi2022motion,jiang2023motiondiffuser} is a key part of the autonomous driving pipeline. Recently, deep learning-based motion prediction methods~\cite{ngiam2021scene,gilles2021home,cui2023gorela} have predicted accurate multimodal distributions of agent behavior. As a result, combining learned motion prediction with planning has lead to stronger performance in autonomous driving~\cite{zeng2019end,sadat2020perceive,casas2021mp3}.

However, current prediction methods focus on prediction accuracy, generally not considering the impact of prediction errors on downstream planning. In the case of adversarial agent behaviors such as jaywalking and red light violations, these errors may be critical for safety. Since adversarial behaviors are rare and often out of distribution, prediction models underestimate the probability of these behaviors, leading to the planner underweighting them.

Due to this underweighting of adversarial behaviors, various safe planning approaches have been developed. These approaches can be classified into two categories: methods that improve planner safety offline before test-time deployment and robust planning methods that promote safety online during the closed-loop planning process. Approaches in the first category use strategies such as training the prediction model to generate adversarial behaviors~\cite{nishimura2023rap,ding2020learning,corso2019adaptive}, training a model to output safe plans~\cite{achiam2017constrained, tessler2018reward,gu2022review,thananjeyan2021recovery,booher2024cimrl,brown2020safe,wang2024coin,hanselmann2022king,wang2021advsim}, and predicting and repairing planner failures~\cite{dawson2023bayesian,zhou2022rocus}. However, these methods all rely on an offline set of adversarial scenarios to improve planner safety, so they may fail to generalize to the multitude of unseen adversarial scenarios. Hence, we instead focus on robust planning methods, which robustify the planner during online test-time deployment, as these methods do not rely on offline adversarial scenarios and hence, do not suffer from this generalization gap.

Robust planning methods can be divided into two categories: risk-sensitive planning~\cite{wang2020game, li2023marc} that evaluates plans according to a conservative risk functional~\cite{whittle1981risk,Pflug2000some} and safety-constrained planning~\cite{hardy2013contingency,zhan2016noncon,thomas2021exact} that enforces safety constraints. Risk-sensitive planners place high weight on high-cost adversarial predictions, leading to robustness against adversarial behaviors. However, conservative risk functionals often do not ensure that some minimum weight is placed on low-cost normal agent behaviors. Hence, risk-sensitive planners may place little to no weight on normal behaviors, leading to overly conservative driving behavior~\cite{trautman2010unfreezing}. On the other hand, safety-constrained planners enforce safety constraints such as probabilistic collision bounds, ensuring that the ego-vehicle follows a plan that is safe with respect to adversarial behaviors. However, these safety constraints may be inflexible, as in they may not work well in all situations.

We propose a robust planning approach that avoids the issues of conservative risk functionals or safety constraints. Instead, we propose to evaluate plans using the risk-neutral expected cost metric, but we compute the expectation with respect to a mixture distribution of normal and adversarial predictions. Via this mixture, we ensure that significant non-zero weight (based on the mixture weights) is placed on both normal and adversarial agent behaviors, leading to a planner that is robust to adversarial behaviors but not overly conservative if agents behave normally. We avoid the issues with safety constraints as we do not enforce them.

We obtain this mixture distribution of normal and adversarial behaviors by using a diffusion motion forecasting model; we use a diffusion model as it can effectively generate realistic predictions~\cite{jiang2023motiondiffuser} and adversarial behaviors~\cite{chang2023controllable, huang2024versatile,xu2023diffscene}. We first train the diffusion model to predict an unbiased distribution of normal agent behaviors. We then generate a distribution of adversarial behaviors by biasing the diffusion model at test time towards predictions that are likely to collide with the plan under consideration. Notably, by biasing the predictions at test time, we can predict unseen adversarial behaviors unlike methods that use offline data of adversarial behaviors and hence, fail to generalize to the multitude of unseen adversarial behaviors. Finally, we evaluate plans using expected cost with respect to a mixture of the normal and adversarial prediction distributions.

Our main contributions are as follows:

\begin{itemize}

\item A robust planning method for autonomous driving that evaluates the expected cost of a candidate plan using a mixture distribution of normal and adversarial agent behaviors. We avoid the overly conservative behavior of risk-sensitive planning by placing significant non-zero weights on both normal and adversarial behaviors. We also avoid inflexible hard safety constraints.

\item A method to obtain the mixture distribution of normal and adversarial predictions using a diffusion motion prediction model, which is biased at test time to sample adversarial behaviors for the candidate plan. Our method does not require offline adversarial behaviors, and it can predict unseen adversarial behaviors.

\item We evaluate our method on multiple scenarios containing adversarial behaviors: single-agent jaywalking, multi-agent jaywalking, and red light violation. The results show that our method is robust to adversarial behaviors while not being overly conservative.

\end{itemize}


\section{Related Work}

\label{sec:related_work}

\textbf{Motion Prediction} Deep learning-based methods for motion forecasting~\cite{gupta2018social,gao2020vectornet,ngiam2021scene,shi2022motion,gilles2021home,jiang2023motiondiffuser,seff2023motionlm,cui2023gorela,salzmann2020trajectron++,chai2019multipath} have become popular due to their ability to predict accurate and multimodal behavior distributions. In particular, some methods focus on diverse prediction~\cite{yuan2019diverse,huang2020diversitygan,yuan2020dlow,bhattacharyya2018accurate,Rhinehart2018r2p2,amirian2019social,chen2022scept,narayanan2021divide,zhao2021tnt}, i.e. ensuring high coverage of the prediction modes. Unlike our method, these methods optimize prediction accuracy, not downstream planner performance. Planning-aware motion predictors aim to align prediction with planning, either via special planning-aware losses~\cite{cui2021lookout,casas2020importance,mcallister2022control,huang2020diversitygan} or by training the prediction model end-to-end with a differentiable planner~\cite{huang2024differentiable,sun2023getdipp,liu2024hybridprediction,huang2024dtpp,karkus2023diffstack}. However, they do not focus on predicting rare adversarial behaviors.

\textbf{Adversarial Motion Prediction} Several approaches have been proposed to tackle the issue of predicting adversarial behaviors. Importance sampling~\cite{luo2019importance,kelly2018scalable} encourages the prediction model to not completely miss rare behaviors, but it does not resolve the issue of the prediction model underestimating the probability of rare adversarial behaviors. Markov chain Monte Carlo (MCMC) approaches~\cite{zhou2022rocus,sinha2020neural,delecki2023model,dawson2023bayesian} sample adversarial predictions using MCMC sampling, but unlike importance sampling methods, they may place increased weight on adversarial predictions~\cite{dawson2023bayesian}. Unlike the prior sampling-based approaches, black-box (gradient-free)~\cite{wang2021advsim,corso2021survey} and gradient-based approaches~\cite{hanselmann2022king} generate adversarial behaviors by optimizing an adversarial objective function. Learning-based approaches~\cite{nishimura2023rap,ding2020learning,corso2019adaptive} train adversarial predictors, potentially using reinforcement learning~\cite{ding2020learning,corso2019adaptive}. While our work uses adversarial predictions, our primary goal is not adversarial prediction but instead closed-loop robust planning. Notably, for robust planning, generating only adversarial predictions is insufficient as the planner needs to consider both normal and adversarial behaviors to avoid becoming overly conservative.

Various approaches~\cite{dawson2023bayesian,hanselmann2022king,zhou2022rocus,wang2021advsim} generate adversarial scenarios offline and then use these scenarios to improve planner safety. However, the planner may not be robust to unseen adversarial scenarios that were not generated offline before test-time deployment. In contrast, we propose to bias a diffusion model to sample adversarial predictions at test-time during closed-loop planning. As the diffusion model can predict unseen adversarial behaviors, our planning method can be robust to unseen adversarial behaviors.

\cite{nishimura2023rap} trains a biased predictor that reduces the problem of risk estimation in risk-sensitive planning to sampling adversarial predictions. However, unlike our method, this method trains the biased predictor on adversarial agent behaviors and hence, it may not generalize to unseen adversarial behaviors. Furthermore, this method is similar to risk-sensitive planning and hence, may lead to overly conservative planning.

\textbf{Robust Planning} Unlike approaches that improve safety using offline adversarial scenarios, leading to a potential failure to generalize to unseen adversarial behaviors, robust planning methods encourage safety during online closed-loop planning. Generally, robust planning can be categorized into two categories: risk-sensitive planning and safety-constrained planning. A risk-sensitive planner uses a conservative risk functional~\cite{majumdar2020should} such as worst-case cost~\cite{zhou2022dynamically,zhou2022long}, entropic risk~\cite{whittle1981risk,wang2020game,chandra2022game} or conditional value-at-risk (CVaR)~\cite{Pflug2000some,zhang2024efficientmpc,li2023marc}. However, these risk functionals may place high-weight on high-cost adversarial behaviors and little to no weight on lower-cost normal behavior modes, leading to overly conservative planning~\cite{trautman2010unfreezing}. Safety-constrained methods use safety constraints such as probabilistic collision and risk bounds~\cite{hardy2013contingency,ren2024safe,thomas2021exact,bhanzaf2018footprint,huang2021planning,bernhard2021risk,zhang2024efficient,mustafa2024racp} and reachability and safety sets~\cite{zhan2016noncon,pan2020safe,nakamura2023online,li2021prediction,muthali2023multi,zhou2024robust,khaitan2021safe,lindemann2023safe}. However, safety constraints are always enforced and hence, generally do not adapt to different scenarios; they may be too strict in some situations and too loose in other situations. In contrast, our method overcomes the issues of overly conservative behavior and inflexible constraints as we instead place significant non-zero weight on both normal and adversarial behaviors via computing expected cost with respect to a mixture of the two types of behaviors.

\textbf{Diffusion Models} Diffusion models~\cite{sohl2015deep, ho2020denoising, song2020denoising, song2020score, nichol2021improved, karras2022elucidating} have shown great success in generating images~\cite{dhariwal2021diffusion,nichol2021glide,ramesh2022hierarchical,saharia2022photorealistic,rombach2022high} and videos~\cite{ho2022imagen,ho2022video,yang2023diffusion,blattmann2023videoldm}. Recently, diffusion models have been applied to motion prediction~\cite{gu2022stochastic,jiang2023motiondiffuser,choi2023dice,chen2023equidiff,rempe2023trace,li2023bcdiff,Mao2023leapfrog,westny2024diffusion}, generating predictions which can be biased at test-time. While we also bias a diffusion model's predictions, unlike these works, we focus on predicting adversarial behaviors for the downstream planning task. In addition, diffusion models have been used to generate realistic driving scenarios~\cite{zhong2023guided,guo2023scenedm,yang2023tsdit,wang2024dragtraffic,zhong2023language,chang2023editing,pronovost2023scenariodiff,chitta2024sledge} with some work focusing on safety-critical scenario generation~\cite{chang2023controllable, huang2024versatile,xu2023diffscene}. We note that safety-critical scenario generation focuses on the controllable generation of individual realistic adversarial scenarios offline. In contrast, our work focuses on the setting of online closed-loop robust planning by evaluating plans using a mixture of normal and adversarial behaviors.

Furthermore, diffusion models have been applied as learned policies~\cite{janner2022planning,wang2022diffusion,ajay2022conditional,hansen2023idql,he2023diffusion,pearce2023imitating}. Recently,~\cite{yang2024diffusion} uses a diffusion model as a planner for autonomous driving, adapting it to different scenario by optimizing scenario-specific rewards. In contrast, our method generates adversarial motion predictions to bias a rule-based planner to be robust. Various works have incorporated safety constraints for learned diffusion model planners~\cite{xiao2023safediffuser,feng2024ltldog,kondo2024cgd}. In contrast, our method uses a rule-based planner and no safety constraints.

\section{Methods}

\label{sec;methods}

\subsection{Problem Definition}

\begin{figure}[t]
\centering
\includegraphics[scale=0.25]{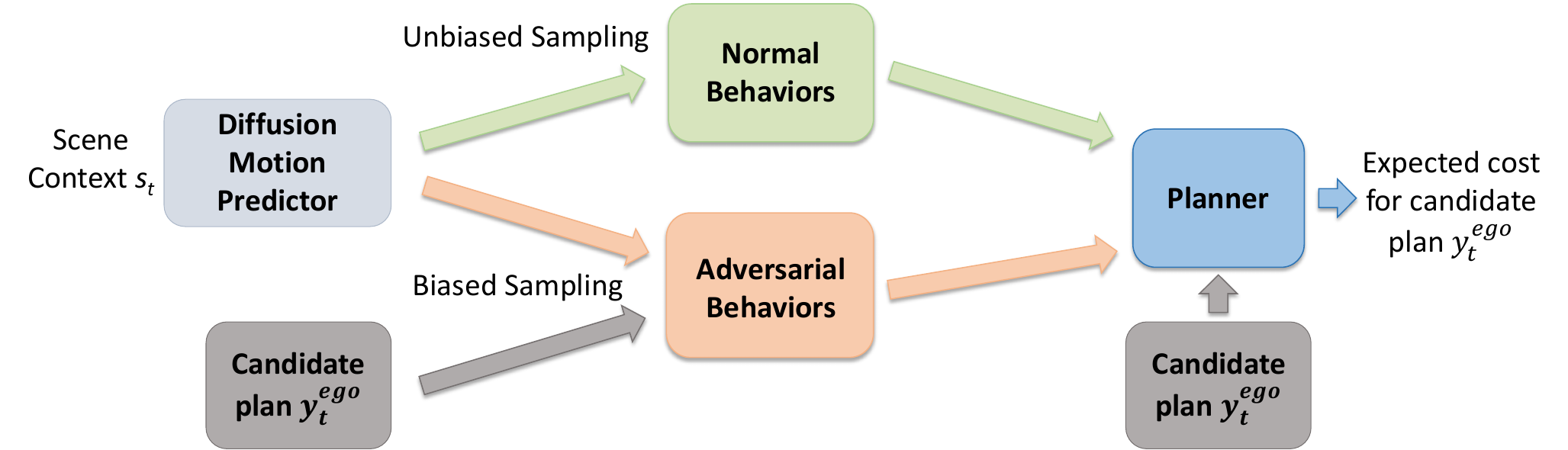}
\caption{Overview of Our Method (best viewed in color at 3x zoom). We first train a diffusion motion prediction model to take in scene context and output normal agent behaviors. We then bias the diffusion motion predictor to predict adversarial agent behaviors for the candidate plan. Finally, the planner computes the expected cost for the candidate plan, using both the normal and adversarial predictions. By taking into account both types of behaviors, our method causes the planner to be robust to adversarial behaviors but not overly conservative.}
\label{fig:method_overview}
\vspace{-5mm}
\end{figure}

We consider the problem of robust planning for driving. In this setting, we do not use a set of offline adversarial scenarios to avoid issues with generalizing to unseen scenarios. Let the cost of a candidate plan (future ego-trajectory) $y^{ego}_t$ given the joint future trajectories $y^{a}_t$ of other agents be given by a cost function $c(y^{ego}_t, y^{a}_t)$. We assume that the planner uses a motion predictor that outputs $p(y^{a}_t|s_t)$, a distribution of future behaviors given the scene context $s_t$; note that $s_t$ includes the history of agents, HD map, etc. Cost is a random variable $C = c(y^{ego}_t, Y^{a}_t)$, where predictions $Y^{a}_t$ are sampled from $p(y^{a}_t|s_t)$.

We formulate the robust planning problem as constrained optimization, where we select the plan $y^{ego}_t$ that minimizes the risk measure $R_p[C]$ from the set $S$ (possibly discrete) of candidate plans subject to optional safety constraints $g_i(y^{ego}_t) \leq h_i, i = 1...G$:
\begin{equation}
\min_{y^{ego}_t \in S} \mathcal{R}_{p}[C] \textrm{ s.t. } g_i(y^{ego}_t) \leq h_i, i = 1...G
\end{equation}
Risk-sensitive planners select $\mathcal{R}_p[C]$ such that $\mathcal{R}_p[C] > \mathbb{E}_p[C]$ while safety constrained planners will enforce constraints $g_i(y^{ego}_t) \leq h_i, i = 1...G$. However, risk-sensitive planners may overweight high-cost adversarial predictions while placing little to no weight on lower-cost normal predictions, leading to overly conservative planning. Safety constrained planners may enforce inflexible constraints that do not work well in all scenarios.

Our method, mixed adversarial diffusion predictions (MAD), shown in Figure~\ref{fig:method_overview}, proposes instead to use expected cost, i.e. we set $\mathcal{R}_p[C] = \mathbb{E}_p[C]$, with respect to a mixture distribution of normal and adversarial agent behaviors. By mixing the adversarial and normal behavior distributions, we ensure that significant non-zero weight is placed on both behavior types during planning, allowing the planner to be robust to adversarial behaviors but not overly conservative. We generate the distributions of normal agent behaviors and adversarial agent behaviors using a diffusion model. We then incorporate the mixture distribution into a state-of-the-art planner, PDM-Closed~\cite{Dauner2023CORL}.

\subsection{Preliminaries: Diffusion Models for Motion Prediction}
\label{sec:diffusion_pred}
We use the MotionDiffuser diffusion motion prediction model~\cite{jiang2023motiondiffuser}, which we briefly describe below. Diffusion motion prediction models, with learned parameters $\boldsymbol{\theta}$, aim to sample from the distribution $p_{\boldsymbol{\theta}}(y^{a}_t|s_t)$ of future agent trajectories $y^{a}_t$ given the scene context $s_t$. Let $p_{\boldsymbol{\theta}}(y^{a}_t|s_t; \sigma)$ be the probability density function obtained by convolving $p_{\boldsymbol{\theta}}(y^{a}_t|s_t)$ with the zero-mean Gaussian distribution with standard deviation $\sigma$. The diffusion model~\cite{karras2022elucidating,jiang2023motiondiffuser} does not directly learn $p_{\boldsymbol{\theta}}(y^{a}_t|s_t)$ but instead learns, for a range of standard deviations $\sigma$, the score function $\nabla_{y^{a}_t} \log p_{\boldsymbol{\theta}}(y^{a}_t|s_t;\sigma)$, which is then used to sample predictions $y^{a}_t$ from $p_{\boldsymbol{\theta}}(y^{a}_t|s_t)$.

To sample a prediction $y^{a}_t \sim p_{\boldsymbol{\theta}}(y^{a}_t|s_t)$ using a trained diffusion model, we consider the reverse diffusion process $y^{a}_t(k) \sim p_{\boldsymbol{\theta}}(y^{a}_t|s_t; \sigma(k))$, where $k$ is the time variable of this process and $\sigma(k)$ specifies the variance schedule with $\sigma(0) = 0$. We then iteratively denoise the noisy prediction $y^{a}_t(k)$ starting from $y^{a}_t(K)$ to obtain the denoised prediction $y^{a}_t = y^{a}_t(0) \sim p_{\boldsymbol{\theta}}(y^{a}_t|s_t)$, where $K$ is the maximum time of the denoising process. Following~\cite{karras2022elucidating,jiang2023motiondiffuser}, we set $\sigma(k) = k$. We note that $y^{a}_t(k)$ satisfies the following ODE:

\vspace{-10pt}
\begin{equation}
\label{eqn:diffusion_ode}
dy^{a}_t(k) = -\dot \sigma(k) \sigma(k) \nabla_{y^{a}_t} \log p_{\boldsymbol{\theta}}(y^{a}_t|s_t;\sigma(k)) dk
\end{equation}
We sample a prediction $y^{a}_t$ using the diffusion model by solving this ODE for $y^{a}_t = y^{a}_t(0)$ using Heun's second order method~\cite{ascher1998computer,karras2022elucidating}. For the initial condition $y^{a}_t(K)$, we assume that $\sigma(K) = \sigma_{\max}$ is large and approximate $p_{\boldsymbol{\theta}}(y^{a}_t|s_t;\sigma(K)) \approx \mathcal{N}(y^{a}_t; \boldsymbol{0}, \sigma_{\max}^2 \boldsymbol{I})$. Hence, we sample $y^{a}_t(K)$ from $\mathcal{N}(y^{a}_t; \boldsymbol{0}, \sigma_{\max}^2 \boldsymbol{I})$.

We train the diffusion model to learn the score function $\nabla_{y^{a}_t} \log p_{\boldsymbol{\theta}}(y^{a}_t|s_t;\sigma)$. Given a sample $(s_t, y^{a}_t)$ from the data distribution $p^*(s_t, y^{a}_t)$, we train a denoiser $D_{\boldsymbol{\theta}}(y^{a}_t; s_t, \sigma)$ to recover the ground truth future trajectories $y^{a}_t$ given noisy trajectories $y^{a}_t + \boldsymbol{\epsilon}$, where $\boldsymbol{\epsilon} \sim \mathcal{N}(\boldsymbol{0}, \sigma^2 \boldsymbol{I})$. The denoiser's training loss is the expected $L_2$ denoising error, where $\sigma$ is sampled from a distribution $q(\sigma)$:
\begin{equation}
\mathbb{E}_{(s_t, y^{a}_t) \sim p^*} \mathbb{E}_ {\sigma \sim q(\sigma)}\mathbb{E}_{\boldsymbol{\epsilon} \sim \mathcal{N}(\boldsymbol{0}, \sigma^2 \boldsymbol{I})} ||D_{\boldsymbol{\theta}}(y^{a}_t + \boldsymbol{\epsilon}; s_t, \sigma) - y^{a}_t||_2^2
\end{equation}

We can then obtain the score function $\nabla_{y^{a}_t} \log p_{\boldsymbol{\theta}}(y^{a}_t|s_t;\sigma)$ using the trained denoiser $D_{\boldsymbol{\theta}}(y^{a}_t; s_t, \sigma)$:
\begin{equation}
\nabla_{y^{a}_t} \log p_{\boldsymbol{\theta}}(y^{a}_t|s_t;\sigma) = (D_{\boldsymbol{\theta}}(y^{a}_t; s_t, \sigma) - y^{a}_t) / \sigma^2
\end{equation}

\subsection{Adversarially Biasing Motion Prediction}

\label{sec:methods_biasing}

We have discussed how to train a diffusion model and how to sample from its distribution $p_{\boldsymbol{\theta}}(y^{a}_t|s_t)$ of normal agent behaviors. However, this diffusion model will underestimate the probability of adversarial agent behaviors, which are rare and out of distribution. Hence, in addition to sampling normal agent predictions, we propose to sample adversarial agent behaviors by guiding the diffusion model towards predictions that are likely to collide with the candidate plan.

Given the diffusion prediction model's distribution $p_{\boldsymbol{\theta}}(y^{a}_t|s_t)$, we follow~\cite{jiang2023motiondiffuser} and bias its predictions using a biasing distribution. However, unlike~\cite{jiang2023motiondiffuser}, we sample biased predictions that are adversarial with regards to a candidate plan $y^{ego}_t$, so we condition the biasing distribution $q(y^{a}_t|s_t, y^{ego}_t)$ on the candidate plan. Consider sampling from the adversarial prediction distribution $p_{\boldsymbol{\theta}, b}(y^{a}_t|s_t, y^{ego}_t) = p_{\boldsymbol{\theta}}(y^{a}_t|s_t) q(y^{a}_t|s_t, y^{ego}_t)$. We can sample from this distribution by replacing the unbiased score function in (\ref{eqn:diffusion_ode}) with the adversarial distribution's score function $\nabla_{y^{a}_t} \log (p_{\boldsymbol{\theta}}(y^{a}_t|s_t; \sigma) q(y^{a}_t|s_t, y^{ego}_t; \sigma)) = \nabla_{y^{a}_t} \log p_{\boldsymbol{\theta}}(y^{a}_t|s_t; \sigma) + \nabla_{y^{a}_t} \log q(y^{a}_t|s_t, y^{ego}_t; \sigma)$. Hence, we can bias the diffusion model's predictions via the biasing score function $\nabla_{y^{a}_t} \log q(y^{a}_t|s_t, y^{ego}_t; \sigma)$. However, we currently have not defined the biasing score function. Following~\cite{jiang2023motiondiffuser}, we approximate the biasing score function $\nabla_{y^{a}_t} \log q(y^{a}_t|s_t, y^{ego}_t; \sigma)$ using a differentiable loss $\mathcal{L}$ evaluated on denoised trajectories $D_{\boldsymbol{\theta}}(y^{a}_t; s_t, \sigma)$, where $\lambda$ is the biasing loss weight:
\vspace{-1mm}
\begin{equation}
\nabla_{y^{a}_t} \log q(y^{a}_t|s_t, y^{ego}_t; \sigma) \approx \lambda \nabla_{y^{a}_t} \mathcal{L}(D_{\boldsymbol{\theta}}(y^{a}_t; s_t, \sigma), y^{ego}_t)
\end{equation}

Now, we only need to define the loss $\mathcal{L}$ to bias the diffusion model to output adversarial predictions with respect to a candidate plan $y^{ego}_t$. We set the loss $\mathcal{L}(y^{a}_t, y^{ego}_t)$ so that low loss corresponds to a low distance between the candidate plan $y^{ego}_t$ and the ``closest" agent's predicted trajectory, encouraging the ego-vehicle and the ``closest" agent to collide.

We define $\mathcal{L}(y^{a}_t, y^{ego}_t)$ formally below. Let $y^{a}_t = \{(s^{a_j}_{t + 1}, ..., s^{a_j}_{t + H})\}_{j=1}^{A_t}$, where $s^{a_j}_t$ is the position of agent $a_j$ at time $t$, $A_t$ is the number of agents (excluding the ego-vehicle) at time $t$, and $H$ is the  horizon. Let the candidate plan $y^{ego}_t = (s^{ego}_{t + 1}, ..., s^{ego}_{t + H})$, where $s^{ego}_t$ is the ego-vehicle position at time $t$. Then, we denote the agent $a_{min}$ to be the agent which achieves the minimum $L_2$ distance between its predicted trajectory and the candidate plan, i.e.
\vspace{-1mm}
\begin{equation}
a_{min} = \argmin_{a_j} \min_{1 \leq h \leq H} ||s^{a_j}_{t + h} - s^{ego}_{t + h}||_2^2.
\end{equation}

We then define $\mathcal{L}(y^{a}_t, y^{ego}_t)$ to be the average $L_1$ distance between the candidate plan and the predicted trajectory of the ``closest" agent $a_{min}$:
\vspace{-1mm}
\begin{equation}
\mathcal{L}(y^{a}_t, y^{ego}_t) = \frac{1}{H} \sum_{h=1}^H \big|\big|s^{a_{min}}_{t + h} - s^{ego}_{t + h}\big|\big|_1
\end{equation}

Using this loss function, we sample future agent predictions $y^{a}_t \sim p_{\boldsymbol{\theta}, b}(y^{a}_t|s_t, y^{ego}_t)$ so that they are closer and hence, likely to collide, with a candidate plan. Predictions from the adversarial distribution are still encouraged to be realistic due to the diffusion model's distribution $p_{\boldsymbol{\theta}}(y^{a}_t|s_t)$.

\subsection{Mixture of Adversarial and Normal Predictions for Planning}

\label{sec:methods_mix}

We then create a mixture distribution that mixes the adversarial $p_{\boldsymbol{\theta}, b}(y^{a}_t|s_t, y^{ego}_t)$ and normal prediction distributions $p_{\boldsymbol{\theta}}(y^{a}_t|s_t)$ and evaluate the expected cost of a candidate plan with respect to this mixture. Letting $w_b$ be the mixture weight of the adversarial distribution $p_{\boldsymbol{\theta}, b}(y^{a}_t|s_t, y^{ego}_t)$, our planner computes the expected cost $\mathbb{E}_{y^{a}_t \sim p_{mix}}[c(y^{ego}_t, y^{a}_t)]$ of candidate ego-plans $y^{ego}_t$ with respect to the mixture distribution $p_{mix}(y^{a}_t|s_t, y^{ego}_t) = (1-w_b) p_{\boldsymbol{\theta}}(y^{a}_t|s_t) + w_b p_{\boldsymbol{\theta}, b}(y^{a}_t|s_t, y^{ego}_t)$.

In practice, we approximate the expected cost via Monte-Carlo sampling by sampling unbiased normal trajectories from $p_{\boldsymbol{\theta}}(y^{a}_t|s_t)$ and adversarial trajectories from $p_{\boldsymbol{\theta}, b}(y^{a}_t|s_t, y^{ego}_t)$ in a ratio of $1 - w_b$ to $w_b$. By computing expected cost using both normal and adverarial agent behaviors during planning, we ensure that our planner is robust to adversarial agent behaviors but not overly conservative.

\subsection{Extending the PDM-Closed Planner}

We build on a state-of-the-art rule-based planner PDM-Closed~\cite{Dauner2023CORL}. PDM-Closed first selects a centerline and proposes a fixed number of candidate plans using Intelligent Driver Model (IDM)~\cite{treiber2000congested} policies with various target speeds and lateral centerline offsets. It then forecasts agents using a constant velocity assumption, simulates the proposed plans, and scores the plans using a hand-defined cost function $c(y^{ego}_t, y^{a}_t)$. This cost function considers factors such as collisions, time to collision, progress, comfort, speed limit, and staying within the drivable area.

PDM-Closed does not consider multiple potential agent behaviors. We extend this planner so that it evaluates expected cost with respect to a mixture of adversarial and normal behaviors.

\subsection{Implementation Details}

We train MotionDiffuser~\cite{jiang2023motiondiffuser} on around 160K scenes from the NuPlan~\cite{nuplan} training set. For each candidate plan, the planner considers 10 sampled predictions, two normal predictions and eight adversarial predictions ($w_b = 0.8$). We set $\lambda$, the weight for the biasing loss, to 0.5, and we clip the biasing score function as proposed by~\cite{jiang2023motiondiffuser}.

\section{Results}

\label{sec:results}

\begin{table*}[]
\vspace{2mm}
\caption{Overall results on all benchmarks.}
\label{tab:results}
\centering
\vspace{-2mm}
\begin{tabular}{c c || c c c}
\toprule
Method & Overall & Single-Agent Jaywalking & Multi-Agent Jaywalking & Red Light Violation \\
\midrule
CV & 66.2 & 68.1 & 66.3 & 64.3 \\
EC & 79.5 & 79.6 & 79.6 & 79.2 \\
CVaR & 83.5 & 84.1 & 79.2 & \textit{87.2} \\
WC & 79.4 & 70.5 & 81.3 & 86.5\\
Col-P 0 & 81.6 & 83.6 & 82.4 & 78.7 \\
Col-P 0.1 & 81.8 & 85.6 & 80.7 & 79.2 \\
MAD & \textbf{86.6} & \textit{87.1} & \textit{85.5} & \textit{87.2} \\
\bottomrule
\end{tabular}
\begin{tablenotes}
Bold indicates best overall method; italics indicate best method(s) on the individual scenario types. Our method MAD achieves the best overall performance over all benchmarks, with \textbf{13.4\%} error, a \textbf{18.8\%} error rate reduction, compared to the second best method CVaR, with \textbf{16.5\%} error.
\end{tablenotes}
\vspace{-5mm}
\end{table*}

\subsection{Evaluation Metric}

We evaluate our method on closed-loop driving scenarios constructed in the NuPlan~\cite{nuplan} simulator. For the evaluation metric, we use NuPlan's closed-loop score (CLS)~\cite{Dauner2023CORL}, which evaluates closed-loop planning on a percentage scale from 0\% to 100\%. At-fault collisions, insufficient progress, and drivable area violations reduce the score to 0. Otherwise, the closed-loop score computes a weighted average of metrics based on time to collision, progress, following the speed limit, driving in the correct direction, and comfort.

\subsection{Evaluation Scenarios}

As the standard NuPlan benchmark, Val14~\cite{Dauner2023CORL}, is saturated (all baselines achieve between 91\% and 92\% CLS), we evaluate on single-agent jaywalking, multi-agent jaywalking, and red light violation scenarios instead due to their focus on adversarial behaviors, a focus that Val14 lacks.

\textbf{Single-Agent Jaywalking Benchmark} The jaywalking pedestrian initially starts at rest on the side of the road. The pedestrian either runs quickly across the road or walks slowly across the road; in both cases, we evaluate with varying accelerations and locations of the jaywalker. If the pedestrian runs quickly across the road, then, the ego-vehicle must slow down for the jayrunning pedestrian to avoid collision. Otherwise, if the pedestrian jaywalks slowly, then the ego-vehicle should speed up slightly as it will pass the pedestrian before the pedestrian enters the ego-vehicle's lane. If the ego-vehicle instead slows down, then, the pedestrian will enter the ego-vehicle's lane, making a collision likely due to not having enough time to slow down to a stop. Hence, this scenario evaluates whether a planner can behave conservatively when necessary (fast jayrunner) but not overly conservative (slow jaywalker).

\textbf{Multi-Agent Jaywalking Benchmark} To examine whether our method MAD can scale to multi-agent scenarios, we evaluate our method in a multi-agent jaywalking benchmark. While the previous jaywalking scenario contains only one agent (excluding the ego-vehicle), in this scenario, we have added several vehicles, which drive non-adversarially, to the single-agent jaywalking scenario.

\textbf{Red Light Violation Benchmark} To demonstrate that our method MAD can effectively handle varying types of scenarios with adversarial agent behaviors, we additionally evaluate in a red light violation scenario. At a given time, a vehicle violates the red light; the ego-vehicle must slow down for this adversarial vehicle. We vary the accelerations and red light violation times of the adversarial vehicle.

\subsection{Baselines}

\label{sec:baselines}

We compare against several baselines for robust planning. All baselines (besides constant velocity) use only unbiased diffusion model predictions from the normal behavior distribution.

\textbf{Constant Velocity (CV)}: This baseline is the PDM-Closed~\cite{Dauner2023CORL} planner, which assumes that agents move at constant velocity.

\textbf{Expected Cost (EC)}: This method is a restriction of our method to the case where the expected cost is computed using only normal predictions of agent behavior. EC is a baseline with low risk-sensitivity.

\textbf{CVaR}: This method is a risk-sensitive planner with a medium level of risk-sensitivity; it uses the CVaR~\cite{Pflug2000some} conservative risk functional. The CVaR risk functional depends on the risk-sensitivity value $r$, where $r \in (0, 1)$ and larger $r$ corresponds to higher risk-sensitivity. We select the intermediate risk-sensitivity value $r = 0.5$ for this baseline. We note that due to the limiting behavior of CVaR, at low risk-sensitivity $r$, CVaR($r$) behaves similar to the risk-neutral expected cost (EC) baseline while at high risk-sensitivity $r$, CVaR($r$) behaves similar to the worst-case (WC) baseline described below. We estimate CVaR using a Monte-Carlo estimator similar to~\cite{trindade2007financial,hong2014montecarlo}.

\textbf{Worst Case (WC)}: This method scores plans by computing the worst-case cost, which is computed by considering the cost with respect to every sampled agent prediction. This conservative planner has high risk-sensitivity as it considers only the most adversarial high-cost prediction.

\textbf{Collision probability (Col-P) 0, 0.1}: These methods use the expected cost metric, but they additionally enforce safety constraints on collision probability. We compute the collision probability using Monte Carlo sampling, i.e. the collision probability for a plan is the proportion of agent predictions that result in a collision. Col-P 0 constrains the collision probability to equal 0 and Col-P 0.1 constrains the collision probability to be no greater than 0.1.

As we focus on robust planning approaches that do not rely on a set of offline adversarial scenarios, we do not compare to approaches~\cite{nishimura2023rap,booher2024cimrl,hanselmann2022king,wang2021advsim} that improve safety using offline adversarial scenarios and hence, potentially fail to generalize to unseen adversarial scenarios.

\subsection{Overall Results}

\label{sec:overall}

\begin{table*}[]
\vspace{2mm}
\caption{Results on the single-agent (left) and multi-agent (right) jaywalking benchmarks.}
\label{tab:results_jaywalk}
\centering
\begin{tabular}{c c || c c}
\toprule
Method & Single-Agent Jaywalking & Fast & Slow \\
\midrule
CV & 68.1 & 53.3 & 82.8 \\
EC & 79.6 & 59.2 & \textit{100.0} \\
CVaR & 84.1 & 68.3 & \textit{100.0} \\
WC & 70.5 & \textit{91.9} & 49.0 \\
Col-P 0 & 83.6 & 74.4 & 92.8 \\
Col-P 0.1 & 85.6 & 71.3 & \textit{100.0} \\
MAD & \textbf{87.1} & 74.2 & \textit{100.0} \\
\bottomrule
\end{tabular}\hspace{2cm}
\begin{tabular}{c c || c c}
\toprule
Method & Multi-Agent Jaywalking & Fast & Slow \\
\midrule
CV & 66.3 & 49.8 & 82.8 \\
EC & 79.6 & 59.2 & \textit{100.0} \\
CVaR & 79.2 & 63.4 & 95.0 \\
WC & 81.3 & \textit{78.2} & 84.4 \\
Col-P 0 & 82.4 & 72.0 & 92.8 \\
Col-P 0.1 & 80.7 & 61.4 & \textit{100.0} \\
MAD & \textbf{85.5} & 76.8 & 94.2 \\
\bottomrule
\end{tabular}
\begin{tablenotes}
Bold indicates best method on the benchmark; italics indicate best method(s) on the sub-benchmarks involving the fast jayrunner and slow jaywalker. Our method MAD achieves the best performance on the single-agent jaywalking scenario, with \textbf{12.9\%} error, a \textbf{10.4\%} error rate reduction, compared to the second-best method Col-P 0.1, with \textbf{14.4\%} error. On the multi-agent jaywalking scenario, our method also achieves the best performance, with \textbf{14.5\%} error, a \textbf{17.6\%} error rate reduction, compared to the next best method Col-P 0, with \textbf{17.6\%} error.
\end{tablenotes}
\vspace{-12pt}
\end{table*}

We present the overall results in Table \ref{tab:results}; we note that our method and all baselines do not use any scenario-specific hyperparameters or models. We observe that our method MAD achieves a \textbf{18.8\%} error rate reduction over the second-best method CVaR. This gain shows that our method performs well in a variety of scenarios with adversarial behaviors; computing expected cost with respect to a mixture of adversarial and normal behaviors allows our method to consider both types of behaviors. Furthermore, we note that the baselines either perform poorly in all scenarios or do not perform consistently well in all scenarios.

\subsection{Single-Agent Jaywalking Benchmark}

\label{sec:jaywalking}

We present the results in Table \ref{tab:results_jaywalk} (left). We observe that our method MAD achieves a \textbf{10.4\%} error rate reduction over the second-best method Col-P 0.1. These gains demonstrate that our method of computing expected cost with respect to a mixture of adversarial behaviors and normal behaviors effectively considers both types of agent behaviors, allowing the planner to react conservatively to the fast jayrunner but behave normally when necessary for the slow jaywalker. We observe that as risk-sensitivity increases, the baselines (EC, CVaR, WC) become more conservative as shown by the results in the fast jayrunning scenario. In particular, the worst-case WC baseline performs poorly in the slow jaywalking scenario, showing that highly risk-sensitive planners over-weight high cost behaviors while placing little to no weight on normal behaviors. We also note that all other methods outperform constant-velocity forecasting (CV), showing the utility of incorporating predictions from our diffusion model.

\subsection{Multi-Agent Jaywalking Benchmark}

The results are in Table \ref{tab:results_jaywalk} (right). We observe that our method MAD performs the best, with a \textbf{17.6\%} error rate reduction over the second best method Col-P 0, showing that MAD effectively scales to scenarios with multiple agents.

\subsection{Red Light Violation Benchmark}

To demonstrate that our method MAD can perform well across varying scenarios with adversarial agent behaviors, we additionally present results in a red light violation scenario in Table \ref{tab:results} (right-most column). Our method MAD performs the best in this scenario, tied with the CVaR baseline. This strong performance in all scenarios (Table \ref{tab:results}) demonstrates that our method can handle adversarial agent behaviors in a variety of situations as it encourages robustness online without relying on an offline set of adversarial scenarios. Furthermore, we observe that the collision probability bound baselines, Col-P 0 and Col-P 0.1, do not improve on the expected cost baseline. We hypothesize that this lack of improvement is due to the collision probability constraints being too ``loose" in this scenario, i.e. the plans that don't satisfy the constraints already have high expected cost, so the constraints do not significantly change the planner's behavior. Hence, the safety constraints enforced by a safety-constrained method may not be appropriate for all scenarios.

\subsection{Ablation Study}

\begin{table}
\caption{Ablation study on the single-agent jaywalking benchmark}
\centering
\begin{tabular}{c c }
\toprule
Method & CLS \\ 
\midrule
$w_b = 0$ (EC) & 79.6 \\ 
$w_b = 0.2$ & 80.5 \\ 
$w_b = 0.4$ & 83.4 \\ 
$w_b = 0.6$ & \textbf{87.1} \\ 
$w_b = 0.8$ (Ours) & \textbf{87.1} \\ 
$w_b = 1.0$ & 62.1 \\ 
\bottomrule
\end{tabular}\hspace{2cm}
\begin{tabular}{c c}
\toprule
Method & CLS \\
\midrule
$\lambda = 0$ (EC) & 79.6 \\
$\lambda = 0.25$ & \textbf{87.1} \\
$\lambda = 0.5$ (Ours) & \textbf{87.1} \\
$\lambda = 1.0$ & \textbf{87.1} \\
\bottomrule
\end{tabular}

\begin{tablenotes}
Our method performs the best. In addition, all methods that use a mixture of normal and adversarial agent behaviors outperform the methods that use only normal agent predictions (EC) or only adversarial predictions.
\end{tablenotes}
\label{tab:ablation}
\vspace{-1mm}
\end{table}

We conduct an ablation study (Table \ref{tab:ablation}) for two hyperparameters of our method: $w_b$, the mixture weight of the adversarial distribution, and $\lambda$, the biasing loss weight when sampling from the adversarial distribution.

We observe that our method ($w_b = 0.8$) performs the best. It outperforms $w_b = 0$, the expected cost baseline that considers only normal agent predictions, and $w_b = 1.0$, which considers only adversarial agent predictions, showing the importance of considering a mixture of normal and adversarial predictions so that our method is robust to adversarial behaviors while not being overly conservative.

We observe when ablating $\lambda$ that any method using a mixture of normal and adversarial agent behaviors (i.e. any method with $\lambda > 0$) outperforms the $\lambda = 0$ (expected cost) baseline, which considers only normal agent behaviors, again showing the importance of considering adversarial agent behaviors. In addition, we note that performance appears to be insensitive to $\lambda$ as long as $\lambda > 0$. This insensitivity may be due to two reasons. First, the diffusion model's distribution encourages adversarial predictions to still be realistic as mentioned in section \ref{sec:methods_biasing}. Second, the cost function used by the planner is bounded. Hence, making the adversarial predictions more adversarial may not change the cost of the plan with respect to the adversarial predictions, leading to little to no change in planning behavior.

\section{Conclusion}

\label{sec:conclusion}

We propose a robust planning method that evaluates candidate plans by computing expected cost with respect to a mixture of normal and adversarial behaviors. We generate the normal behavior distribution by sampling predictions from a diffusion prediction model and generate the adversarial distribution by biasing the diffusion model to sample predictions that are likely to collide with the candidate plan. Our experiments demonstrate the effectiveness of our method.





\bibliographystyle{IEEEtran}
\bibliography{IEEEabrv, camera_ready}  

\begin{thebibliography}{100}
\providecommand{\url}[1]{#1}
\csname url@rmstyle\endcsname
\providecommand{\newblock}{\relax}
\providecommand{\bibinfo}[2]{#2}
\providecommand\BIBentrySTDinterwordspacing{\spaceskip=0pt\relax}
\providecommand\BIBentryALTinterwordstretchfactor{4}
\providecommand\BIBentryALTinterwordspacing{\spaceskip=\fontdimen2\font plus
\BIBentryALTinterwordstretchfactor\fontdimen3\font minus \fontdimen4\font\relax}
\providecommand\BIBforeignlanguage[2]{{%
\expandafter\ifx\csname l@#1\endcsname\relax
\typeout{** WARNING: IEEEtran.bst: No hyphenation pattern has been}%
\typeout{** loaded for the language `#1'. Using the pattern for}%
\typeout{** the default language instead.}%
\else
\language=\csname l@#1\endcsname
\fi
#2}}

\bibitem{gao2020vectornet}
J.~Gao, C.~Sun, H.~Zhao, Y.~Shen, D.~Anguelov, C.~Li, and C.~Schmid, ``Vectornet: Encoding hd maps and agent dynamics from vectorized representation,'' in \emph{Proceedings of the IEEE/CVF Conference on Computer Vision and Pattern Recognition}, 2020, pp. 11\,522--11\,530.

\bibitem{salzmann2020trajectron++}
T.~Salzmann, B.~Ivanovic, P.~Chakravarty, and M.~Pavone, ``Trajectron++: Dynamically-feasible trajectory forecasting with heterogeneous data,'' in \emph{Computer Vision--ECCV 2020: 16th European Conference, Glasgow, UK, August 23--28, 2020, Proceedings, Part XVIII 16}.\hskip 1em plus 0.5em minus 0.4em\relax Springer, 2020, pp. 683--700.

\bibitem{shi2022motion}
S.~Shi, L.~Jiang, D.~Dai, and B.~Schiele, ``Motion transformer with global intention localization and local movement refinement,'' \emph{Advances in Neural Information Processing Systems}, vol.~35, pp. 6531--6543, 2022.

\bibitem{jiang2023motiondiffuser}
C.~Jiang, A.~Cornman, C.~Park, B.~Sapp, Y.~Zhou, D.~Anguelov, \emph{et~al.}, ``Motiondiffuser: Controllable multi-agent motion prediction using diffusion,'' in \emph{Proceedings of the IEEE/CVF Conference on Computer Vision and Pattern Recognition}, 2023, pp. 9644--9653.

\bibitem{ngiam2021scene}
J.~Ngiam, B.~Caine, V.~Vasudevan, Z.~Zhang, H.-T.~L. Chiang, J.~Ling, R.~Roelofs, A.~Bewley, C.~Liu, A.~Venugopal, \emph{et~al.}, ``Scene transformer: A unified architecture for predicting multiple agent trajectories,'' \emph{arXiv preprint arXiv:2106.08417}, 2021.

\bibitem{gilles2021home}
T.~Gilles, S.~Sabatini, D.~Tsishkou, B.~Stanciulescu, and F.~Moutarde, ``Home: Heatmap output for future motion estimation,'' in \emph{2021 IEEE International Intelligent Transportation Systems Conference (ITSC)}.\hskip 1em plus 0.5em minus 0.4em\relax IEEE, 2021, pp. 500--507.

\bibitem{cui2023gorela}
A.~Cui, S.~Casas, K.~Wong, S.~Suo, and R.~Urtasun, ``Gorela: Go relative for viewpoint-invariant motion forecasting,'' in \emph{2023 IEEE International Conference on Robotics and Automation (ICRA)}.\hskip 1em plus 0.5em minus 0.4em\relax IEEE, 2023, pp. 7801--7807.

\bibitem{zeng2019end}
W.~Zeng, W.~Luo, S.~Suo, A.~Sadat, B.~Yang, S.~Casas, and R.~Urtasun, ``End-to-end interpretable neural motion planner,'' in \emph{Proceedings of the IEEE/CVF Conference on Computer Vision and Pattern Recognition}, 2019, pp. 8660--8669.

\bibitem{sadat2020perceive}
A.~Sadat, S.~Casas, M.~Ren, X.~Wu, P.~Dhawan, and R.~Urtasun, ``Perceive, predict, and plan: Safe motion planning through interpretable semantic representations,'' in \emph{Computer Vision--ECCV 2020: 16th European Conference, Glasgow, UK, August 23--28, 2020, Proceedings, Part XXIII 16}.\hskip 1em plus 0.5em minus 0.4em\relax Springer, 2020, pp. 414--430.

\bibitem{casas2021mp3}
S.~Casas, A.~Sadat, and R.~Urtasun, ``Mp3: A unified model to map, perceive, predict and plan,'' in \emph{Proceedings of the IEEE/CVF Conference on Computer Vision and Pattern Recognition}, 2021, pp. 14\,398--14\,407.

\bibitem{nishimura2023rap}
H.~Nishimura, J.~Mercat, B.~Wulfe, R.~T. McAllister, and A.~Gaidon, ``Rap: Risk-aware prediction for robust planning,'' in \emph{Conference on Robot Learning}.\hskip 1em plus 0.5em minus 0.4em\relax PMLR, 2023, pp. 381--392.

\bibitem{ding2020learning}
W.~Ding, B.~Chen, M.~Xu, and D.~Zhao, ``Learning to collide: An adaptive safety-critical scenarios generating method,'' in \emph{2020 IEEE/RSJ International Conference on Intelligent Robots and Systems (IROS)}.\hskip 1em plus 0.5em minus 0.4em\relax IEEE, 2020, pp. 2243--2250.

\bibitem{corso2019adaptive}
A.~Corso, P.~Du, K.~Driggs-Campbell, and M.~J. Kochenderfer, ``Adaptive stress testing with reward augmentation for autonomous vehicle validatio,'' in \emph{2019 IEEE Intelligent Transportation Systems Conference (ITSC)}.\hskip 1em plus 0.5em minus 0.4em\relax IEEE, 2019, pp. 163--168.

\bibitem{achiam2017constrained}
J.~Achiam, D.~Held, A.~Tamar, and P.~Abbeel, ``Constrained policy optimization,'' in \emph{International conference on machine learning}.\hskip 1em plus 0.5em minus 0.4em\relax PMLR, 2017, pp. 22--31.

\bibitem{tessler2018reward}
C.~Tessler, D.~J. Mankowitz, and S.~Mannor, ``Reward constrained policy optimization,'' \emph{arXiv preprint arXiv:1805.11074}, 2018.

\bibitem{gu2022review}
S.~Gu, L.~Yang, Y.~Du, G.~Chen, F.~Walter, J.~Wang, and A.~Knoll, ``A review of safe reinforcement learning: Methods, theory and applications,'' \emph{arXiv preprint arXiv:2205.10330}, 2022.

\bibitem{thananjeyan2021recovery}
B.~Thananjeyan, A.~Balakrishna, S.~Nair, M.~Luo, K.~Srinivasan, M.~Hwang, J.~E. Gonzalez, J.~Ibarz, C.~Finn, and K.~Goldberg, ``Recovery rl: Safe reinforcement learning with learned recovery zones,'' \emph{IEEE Robotics and Automation Letters}, vol.~6, no.~3, pp. 4915--4922, 2021.

\bibitem{booher2024cimrl}
J.~Booher, K.~Rohanimanesh, J.~Xu, and A.~Petiushko, ``Cimrl: Combining imitation and reinforcement learning for safe autonomous driving,'' \emph{arXiv preprint arXiv:2406.08878}, 2024.

\bibitem{brown2020safe}
D.~Brown, R.~Coleman, R.~Srinivasan, and S.~Niekum, ``Safe imitation learning via fast bayesian reward inference from preferences,'' in \emph{International Conference on Machine Learning}.\hskip 1em plus 0.5em minus 0.4em\relax PMLR, 2020, pp. 1165--1177.

\bibitem{wang2024coin}
L.~Wang, M.~Das, F.~Yang, C.~Duo, B.~Qiao, H.~Dong, S.~Qin, C.~Bansal, Q.~Lin, S.~Rajmohan, \emph{et~al.}, ``Coin: Chance-constrained imitation learning for uncertainty-aware adaptive resource oversubscription policy,'' \emph{arXiv preprint arXiv:2401.07051}, 2024.

\bibitem{hanselmann2022king}
N.~Hanselmann, K.~Renz, K.~Chitta, A.~Bhattacharyya, and A.~Geiger, ``King: Generating safety-critical driving scenarios for robust imitation via kinematics gradients,'' in \emph{European Conference on Computer Vision}.\hskip 1em plus 0.5em minus 0.4em\relax Springer, 2022, pp. 335--352.

\bibitem{wang2021advsim}
J.~Wang, A.~Pun, J.~Tu, S.~Manivasagam, A.~Sadat, S.~Casas, M.~Ren, and R.~Urtasun, ``Advsim: Generating safety-critical scenarios for self-driving vehicles,'' in \emph{Proceedings of the IEEE/CVF Conference on Computer Vision and Pattern Recognition}, 2021, pp. 9904--9913.

\bibitem{dawson2023bayesian}
C.~Dawson and C.~Fan, ``A bayesian approach to breaking things: efficiently predicting and repairing failure modes via sampling,'' \emph{arXiv preprint arXiv:2309.08052}, 2023.

\bibitem{zhou2022rocus}
Y.~Zhou, S.~Booth, N.~Figueroa, and J.~Shah, ``Rocus: Robot controller understanding via sampling,'' in \emph{Conference on Robot Learning}.\hskip 1em plus 0.5em minus 0.4em\relax PMLR, 2022, pp. 850--860.

\bibitem{wang2020game}
M.~Wang, N.~Mehr, A.~Gaidon, and M.~Schwager, ``Game-theoretic planning for risk-aware interactive agents,'' in \emph{2020 IEEE/RSJ International Conference on Intelligent Robots and Systems (IROS)}, 2020, pp. 6998--7005.

\bibitem{li2023marc}
T.~Li, L.~Zhang, S.~Liu, and S.~Shen, ``Marc: Multipolicy and risk-aware contingency planning for autonomous driving,'' \emph{IEEE Robotics and Automation Letters}, 2023.

\bibitem{whittle1981risk}
\BIBentryALTinterwordspacing
P.~Whittle, ``Risk-sensitive linear/quadratic/gaussian control,'' \emph{Advances in Applied Probability}, vol.~13, no.~4, pp. 764--777, 1981. [Online]. Available: \url{http://www.jstor.org/stable/1426972}
\BIBentrySTDinterwordspacing

\bibitem{Pflug2000some}
G.~C. Pflug, \emph{Some Remarks on the Value-at-Risk and the Conditional Value-at-Risk}.\hskip 1em plus 0.5em minus 0.4em\relax Boston, MA: Springer US, 2000, pp. 272--281.

\bibitem{hardy2013contingency}
J.~Hardy and M.~Campbell, ``Contingency planning over probabilistic obstacle predictions for autonomous road vehicles,'' \emph{IEEE Transactions on Robotics}, vol.~29, no.~4, pp. 913--929, 2013.

\bibitem{zhan2016noncon}
W.~Zhan, C.~Liu, C.-Y. Chan, and M.~Tomizuka, ``A non-conservatively defensive strategy for urban autonomous driving,'' in \emph{2016 IEEE 19th International Conference on Intelligent Transportation Systems (ITSC)}, 2016, pp. 459--464.

\bibitem{thomas2021exact}
A.~Thomas, F.~Mastrogiovanni, and M.~Baglietto, ``Exact and bounded collision probability for motion planning under gaussian uncertainty,'' \emph{IEEE Robotics and Automation Letters}, vol.~7, no.~1, pp. 167--174, 2022.

\bibitem{trautman2010unfreezing}
P.~Trautman and A.~Krause, ``Unfreezing the robot: Navigation in dense, interacting crowds,'' in \emph{2010 IEEE/RSJ International Conference on Intelligent Robots and Systems}, 2010, pp. 797--803.

\bibitem{chang2023controllable}
W.-J. Chang, F.~Pittaluga, M.~Tomizuka, W.~Zhan, and M.~Chandraker, ``Controllable safety-critical closed-loop traffic simulation via guided diffusion,'' \emph{arXiv preprint arXiv:2401.00391}, 2023.

\bibitem{huang2024versatile}
Z.~Huang, Z.~Zhang, A.~Vaidya, Y.~Chen, C.~Lv, and J.~F. Fisac, ``Versatile scene-consistent traffic scenario generation as optimization with diffusion,'' \emph{arXiv preprint arXiv:2404.02524}, 2024.

\bibitem{xu2023diffscene}
\BIBentryALTinterwordspacing
C.~Xu, D.~Zhao, A.~Sangiovanni-Vincentelli, and B.~Li, ``Diffscene: Diffusion-based safety-critical scenario generation for autonomous vehicles,'' in \emph{The Second Workshop on New Frontiers in Adversarial Machine Learning}, 2023. [Online]. Available: \url{https://openreview.net/forum?id=hclEbdHida}
\BIBentrySTDinterwordspacing

\bibitem{gupta2018social}
A.~Gupta, J.~Johnson, L.~Fei-Fei, S.~Savarese, and A.~Alahi, ``Social gan: Socially acceptable trajectories with generative adversarial networks,'' in \emph{Proceedings of the IEEE conference on computer vision and pattern recognition}, 2018, pp. 2255--2264.

\bibitem{seff2023motionlm}
A.~Seff, B.~Cera, D.~Chen, M.~Ng, A.~Zhou, N.~Nayakanti, K.~S. Refaat, R.~Al-Rfou, and B.~Sapp, ``Motionlm: Multi-agent motion forecasting as language modeling,'' in \emph{Proceedings of the IEEE/CVF International Conference on Computer Vision}, 2023, pp. 8545--8556.

\bibitem{chai2019multipath}
Y.~Chai, B.~Sapp, M.~Bansal, and D.~Anguelov, ``Multipath: Multiple probabilistic anchor trajectory hypotheses for behavior prediction,'' \emph{arXiv preprint arXiv:1910.05449}, 2019.

\bibitem{yuan2019diverse}
Y.~Yuan and K.~Kitani, ``Diverse trajectory forecasting with determinantal point processes,'' \emph{arXiv preprint arXiv:1907.04967}, 2019.

\bibitem{huang2020diversitygan}
X.~Huang, S.~G. McGill, J.~A. DeCastro, L.~Fletcher, J.~J. Leonard, B.~C. Williams, and G.~Rosman, ``Diversitygan: Diversity-aware vehicle motion prediction via latent semantic sampling,'' \emph{IEEE Robotics and Automation Letters}, vol.~5, no.~4, pp. 5089--5096, 2020.

\bibitem{yuan2020dlow}
Y.~Yuan and K.~Kitani, ``Dlow: Diversifying latent flows for diverse human motion prediction,'' in \emph{Computer Vision--ECCV 2020: 16th European Conference, Glasgow, UK, August 23--28, 2020, Proceedings, Part IX 16}.\hskip 1em plus 0.5em minus 0.4em\relax Springer, 2020, pp. 346--364.

\bibitem{bhattacharyya2018accurate}
A.~Bhattacharyya, B.~Schiele, and M.~Fritz, ``Accurate and diverse sampling of sequences based on a “best of many” sample objective,'' in \emph{Proceedings of the IEEE Conference on Computer Vision and Pattern Recognition}, 2018, pp. 8485--8493.

\bibitem{Rhinehart2018r2p2}
N.~Rhinehart, K.~M. Kitani, and P.~Vernaza, ``R2p2: A reparameterized pushforward policy for diverse, precise generative path forecasting,'' in \emph{Proceedings of the European Conference on Computer Vision (ECCV)}, September 2018.

\bibitem{amirian2019social}
J.~Amirian, J.-B. Hayet, and J.~Pettr{\'e}, ``Social ways: Learning multi-modal distributions of pedestrian trajectories with gans,'' in \emph{Proceedings of the IEEE/CVF Conference on Computer Vision and Pattern Recognition Workshops}, 2019, pp. 0--0.

\bibitem{chen2022scept}
Y.~Chen, B.~Ivanovic, and M.~Pavone, ``Scept: Scene-consistent, policy-based trajectory predictions for planning,'' in \emph{Proceedings of the IEEE/CVF Conference on Computer Vision and Pattern Recognition}, 2022, pp. 17\,082--17\,091.

\bibitem{narayanan2021divide}
S.~Narayanan, R.~Moslemi, F.~Pittaluga, B.~Liu, and M.~Chandraker, ``Divide-and-conquer for lane-aware diverse trajectory prediction,'' in \emph{Proceedings of the IEEE/CVF Conference on Computer Vision and Pattern Recognition}, 2021, pp. 15\,799--15\,808.

\bibitem{zhao2021tnt}
H.~Zhao, J.~Gao, T.~Lan, C.~Sun, B.~Sapp, B.~Varadarajan, Y.~Shen, Y.~Shen, Y.~Chai, C.~Schmid, \emph{et~al.}, ``Tnt: Target-driven trajectory prediction,'' in \emph{Conference on Robot Learning}.\hskip 1em plus 0.5em minus 0.4em\relax PMLR, 2021, pp. 895--904.

\bibitem{cui2021lookout}
A.~Cui, S.~Casas, A.~Sadat, R.~Liao, and R.~Urtasun, ``Lookout: Diverse multi-future prediction and planning for self-driving,'' in \emph{Proceedings of the IEEE/CVF International Conference on Computer Vision}, 2021, pp. 16\,087--16\,096.

\bibitem{casas2020importance}
S.~Casas, C.~Gulino, S.~Suo, and R.~Urtasun, ``The importance of prior knowledge in precise multimodal prediction,'' in \emph{2020 IEEE/RSJ International Conference on Intelligent Robots and Systems (IROS)}, 2020, pp. 2295--2302.

\bibitem{mcallister2022control}
R.~McAllister, B.~Wulfe, J.~Mercat, L.~Ellis, S.~Levine, and A.~Gaidon, ``Control-aware prediction objectives for autonomous driving,'' in \emph{2022 International Conference on Robotics and Automation (ICRA)}.\hskip 1em plus 0.5em minus 0.4em\relax IEEE, 2022, pp. 01--08.

\bibitem{huang2024differentiable}
Z.~Huang, H.~Liu, J.~Wu, and C.~Lv, ``Differentiable integrated motion prediction and planning with learnable cost function for autonomous driving,'' \emph{IEEE Transactions on Neural Networks and Learning Systems}, vol.~35, no.~11, pp. 15\,222--15\,236, 2024.

\bibitem{sun2023getdipp}
J.~Sun, C.~Yuan, S.~Sun, Z.~Liu, T.~Goh, A.~Wong, K.~P. Tee, and M.~H. Ang, ``Get-dipp: Graph-embedded transformer for differentiable integrated prediction and planning,'' in \emph{2023 3rd International Conference on Computer, Control and Robotics (ICCCR)}, 2023, pp. 414--421.

\bibitem{liu2024hybridprediction}
H.~Liu, Z.~Huang, W.~Huang, H.~Yang, X.~Mo, and C.~Lv, ``Hybrid-prediction integrated planning for autonomous driving,'' 2024.

\bibitem{huang2024dtpp}
Z.~Huang, P.~Karkus, B.~Ivanovic, Y.~Chen, M.~Pavone, and C.~Lv, ``Dtpp: Differentiable joint conditional prediction and cost evaluation for tree policy planning in autonomous driving,'' 2024.

\bibitem{karkus2023diffstack}
P.~Karkus, B.~Ivanovic, S.~Mannor, and M.~Pavone, ``Diffstack: A differentiable and modular control stack for autonomous vehicles,'' in \emph{Conference on robot learning}.\hskip 1em plus 0.5em minus 0.4em\relax PMLR, 2023, pp. 2170--2180.

\bibitem{luo2019importance}
Y.~Luo, H.~Bai, D.~Hsu, and W.~S. Lee, ``Importance sampling for online planning under uncertainty,'' \emph{The International Journal of Robotics Research}, vol.~38, no. 2-3, pp. 162--181, 2019.

\bibitem{kelly2018scalable}
M.~O\textquotesingle~Kelly, A.~Sinha, H.~Namkoong, R.~Tedrake, and J.~C. Duchi, ``Scalable end-to-end autonomous vehicle testing via rare-event simulation,'' in \emph{Advances in Neural Information Processing Systems}, S.~Bengio, H.~Wallach, H.~Larochelle, K.~Grauman, N.~Cesa-Bianchi, and R.~Garnett, Eds., vol.~31.\hskip 1em plus 0.5em minus 0.4em\relax Curran Associates, Inc., 2018.

\bibitem{sinha2020neural}
A.~Sinha, M.~O'Kelly, R.~Tedrake, and J.~C. Duchi, ``Neural bridge sampling for evaluating safety-critical autonomous systems,'' \emph{Advances in Neural Information Processing Systems}, vol.~33, pp. 6402--6416, 2020.

\bibitem{delecki2023model}
H.~Delecki, A.~Corso, and M.~Kochenderfer, ``Model-based validation as probabilistic inference,'' in \emph{Learning for Dynamics and Control Conference}.\hskip 1em plus 0.5em minus 0.4em\relax PMLR, 2023, pp. 825--837.

\bibitem{corso2021survey}
A.~Corso, R.~Moss, M.~Koren, R.~Lee, and M.~Kochenderfer, ``A survey of algorithms for black-box safety validation of cyber-physical systems,'' \emph{Journal of Artificial Intelligence Research}, vol.~72, pp. 377--428, 2021.

\bibitem{majumdar2020should}
A.~Majumdar and M.~Pavone, ``How should a robot assess risk? towards an axiomatic theory of risk in robotics,'' in \emph{Robotics Research: The 18th International Symposium ISRR}.\hskip 1em plus 0.5em minus 0.4em\relax Springer, 2020, pp. 75--84.

\bibitem{zhou2022dynamically}
W.~Zhou, Z.~Cao, N.~Deng, X.~Liu, K.~Jiang, and D.~Yang, ``Dynamically conservative self-driving planner for long-tail cases,'' \emph{{IEEE} Trans. Intell. Transport. Syst.}, vol.~24, no.~3, pp. 3476--3488, 2023.

\bibitem{zhou2022long}
W.~Zhou, Z.~Cao, Y.~Xu, N.~Deng, X.~Liu, K.~Jiang, and D.~Yang, ``Long-tail prediction uncertainty aware trajectory planning for self-driving vehicles,'' in \emph{2022 IEEE 25th International Conference on Intelligent Transportation Systems (ITSC)}.\hskip 1em plus 0.5em minus 0.4em\relax IEEE, 2022, pp. 1275--1282.

\bibitem{chandra2022game}
R.~Chandra, M.~Wang, M.~Schwager, and D.~Manocha, ``Game-theoretic planning for autonomous driving among risk-aware human drivers,'' in \emph{2022 International Conference on Robotics and Automation (ICRA)}.\hskip 1em plus 0.5em minus 0.4em\relax IEEE, 2022, pp. 2876--2883.

\bibitem{zhang2024efficientmpc}
L.~Zhang, G.~Pantazis, S.~Han, and S.~Grammatico, ``An efficient risk-aware branch mpc for automated driving that is robust to uncertain vehicle behaviors,'' \emph{arXiv preprint arXiv:2403.18695}, 2024.

\bibitem{ren2024safe}
K.~Ren, C.~Chen, H.~Sung, H.~Ahn, I.~Mitchell, and M.~Kamgarpour, ``Safe chance-constrained model predictive control under gaussian mixture model uncertainty,'' \emph{arXiv preprint arXiv:2401.03799}, 2024.

\bibitem{bhanzaf2018footprint}
H.~Banzhaf, M.~Dolgov, J.~Stellet, and J.~M. Zöllner, ``From footprints to beliefprints: Motion planning under uncertainty for maneuvering automated vehicles in dense scenarios,'' in \emph{2018 21st International Conference on Intelligent Transportation Systems (ITSC)}, 2018, pp. 1680--1687.

\bibitem{huang2021planning}
H.-J. Huang, K.-C. Huang, M.~{\v{C}}{\'a}p, Y.~Zhao, Y.~N. Wu, and C.~L. Baker, ``Planning on a (risk) budget: Safe non-conservative planning in probabilistic dynamic environments,'' in \emph{2021 IEEE International Conference on Robotics and Automation (ICRA)}.\hskip 1em plus 0.5em minus 0.4em\relax IEEE, 2021, pp. 10\,257--10\,263.

\bibitem{bernhard2021risk}
J.~Bernhard and A.~Knoll, ``Risk-constrained interactive safety under behavior uncertainty for autonomous driving,'' in \emph{2021 IEEE Intelligent Vehicles Symposium (IV)}.\hskip 1em plus 0.5em minus 0.4em\relax IEEE, 2021, pp. 63--70.

\bibitem{zhang2024efficient}
C.~Zhang, X.~Wu, J.~Wang, and M.~Song, ``Efficient uncertainty-aware collision avoidance for autonomous driving using convolutions,'' \emph{{IEEE} Trans. Intell. Transport. Syst.}, vol.~25, no.~10, pp. 13\,805--13\,819, 2024.

\bibitem{mustafa2024racp}
K.~A. Mustafa, D.~J. Ornia, J.~Kober, and J.~Alonso-Mora, ``Racp: Risk-aware contingency planning with multi-modal predictions,'' \emph{IEEE Transactions on Intelligent Vehicles}, pp. 1--16, 2024.

\bibitem{pan2020safe}
Y.~Pan, Q.~Lin, H.~Shah, and J.~M. Dolan, ``Safe planning for self-driving via adaptive constrained ilqr,'' in \emph{2020 IEEE/RSJ International Conference on Intelligent Robots and Systems (IROS)}.\hskip 1em plus 0.5em minus 0.4em\relax IEEE, 2020, pp. 2377--2383.

\bibitem{nakamura2023online}
K.~Nakamura and S.~Bansal, ``Online update of safety assurances using confidence-based predictions,'' in \emph{2023 IEEE International Conference on Robotics and Automation (ICRA)}.\hskip 1em plus 0.5em minus 0.4em\relax IEEE, 2023, pp. 12\,765--12\,771.

\bibitem{li2021prediction}
A.~Li, L.~Sun, W.~Zhan, M.~Tomizuka, and M.~Chen, ``Prediction-based reachability for collision avoidance in autonomous driving,'' in \emph{2021 IEEE International Conference on Robotics and Automation (ICRA)}.\hskip 1em plus 0.5em minus 0.4em\relax IEEE, 2021, pp. 7908--7914.

\bibitem{muthali2023multi}
A.~Muthali, H.~Shen, S.~Deglurkar, M.~H. Lim, R.~Roelofs, A.~Faust, and C.~Tomlin, ``Multi-agent reachability calibration with conformal prediction,'' in \emph{2023 62nd IEEE Conference on Decision and Control (CDC)}.\hskip 1em plus 0.5em minus 0.4em\relax IEEE, 2023, pp. 6596--6603.

\bibitem{zhou2024robust}
J.~Zhou, Y.~Gao, O.~Johansson, B.~Olofsson, and E.~Frisk, ``Robust predictive motion planning by learning obstacle uncertainty,'' \emph{arXiv preprint arXiv:2403.06222}, 2024.

\bibitem{khaitan2021safe}
S.~Khaitan, Q.~Lin, and J.~M. Dolan, ``Safe planning and control under uncertainty for self-driving,'' \emph{{IEEE} Trans. Veh. Technol.}, vol.~70, no.~10, pp. 9826--9837, 2021.

\bibitem{lindemann2023safe}
L.~Lindemann, M.~Cleaveland, G.~Shim, and G.~J. Pappas, ``Safe planning in dynamic environments using conformal prediction,'' \emph{IEEE Robotics and Automation Letters}, 2023.

\bibitem{sohl2015deep}
J.~Sohl-Dickstein, E.~Weiss, N.~Maheswaranathan, and S.~Ganguli, ``Deep unsupervised learning using nonequilibrium thermodynamics,'' in \emph{International conference on machine learning}.\hskip 1em plus 0.5em minus 0.4em\relax PMLR, 2015, pp. 2256--2265.

\bibitem{ho2020denoising}
J.~Ho, A.~Jain, and P.~Abbeel, ``Denoising diffusion probabilistic models,'' \emph{Advances in neural information processing systems}, vol.~33, pp. 6840--6851, 2020.

\bibitem{song2020denoising}
J.~Song, C.~Meng, and S.~Ermon, ``Denoising diffusion implicit models,'' \emph{arXiv preprint arXiv:2010.02502}, 2020.

\bibitem{song2020score}
Y.~Song, J.~Sohl-Dickstein, D.~P. Kingma, A.~Kumar, S.~Ermon, and B.~Poole, ``Score-based generative modeling through stochastic differential equations,'' \emph{arXiv preprint arXiv:2011.13456}, 2020.

\bibitem{nichol2021improved}
A.~Q. Nichol and P.~Dhariwal, ``Improved denoising diffusion probabilistic models,'' in \emph{International Conference on Machine Learning}.\hskip 1em plus 0.5em minus 0.4em\relax PMLR, 2021, pp. 8162--8171.

\bibitem{karras2022elucidating}
T.~Karras, M.~Aittala, T.~Aila, and S.~Laine, ``Elucidating the design space of diffusion-based generative models,'' \emph{Advances in Neural Information Processing Systems}, vol.~35, pp. 26\,565--26\,577, 2022.

\bibitem{dhariwal2021diffusion}
P.~Dhariwal and A.~Nichol, ``Diffusion models beat gans on image synthesis,'' \emph{Advances in neural information processing systems}, vol.~34, pp. 8780--8794, 2021.

\bibitem{nichol2021glide}
A.~Nichol, P.~Dhariwal, A.~Ramesh, P.~Shyam, P.~Mishkin, B.~McGrew, I.~Sutskever, and M.~Chen, ``Glide: Towards photorealistic image generation and editing with text-guided diffusion models,'' \emph{arXiv preprint arXiv:2112.10741}, 2021.

\bibitem{ramesh2022hierarchical}
A.~Ramesh, P.~Dhariwal, A.~Nichol, C.~Chu, and M.~Chen, ``Hierarchical text-conditional image generation with clip latents,'' \emph{arXiv preprint arXiv:2204.06125}, vol.~1, no.~2, p.~3, 2022.

\bibitem{saharia2022photorealistic}
C.~Saharia, W.~Chan, S.~Saxena, L.~Li, J.~Whang, E.~L. Denton, K.~Ghasemipour, R.~Gontijo~Lopes, B.~Karagol~Ayan, T.~Salimans, \emph{et~al.}, ``Photorealistic text-to-image diffusion models with deep language understanding,'' \emph{Advances in Neural Information Processing Systems}, vol.~35, pp. 36\,479--36\,494, 2022.

\bibitem{rombach2022high}
R.~Rombach, A.~Blattmann, D.~Lorenz, P.~Esser, and B.~Ommer, ``High-resolution image synthesis with latent diffusion models,'' in \emph{Proceedings of the IEEE/CVF conference on computer vision and pattern recognition}, 2022, pp. 10\,674--10\,685x.

\bibitem{ho2022imagen}
J.~Ho, W.~Chan, C.~Saharia, J.~Whang, R.~Gao, A.~Gritsenko, D.~P. Kingma, B.~Poole, M.~Norouzi, D.~J. Fleet, \emph{et~al.}, ``Imagen video: High definition video generation with diffusion models,'' \emph{arXiv preprint arXiv:2210.02303}, 2022.

\bibitem{ho2022video}
J.~Ho, T.~Salimans, A.~Gritsenko, W.~Chan, M.~Norouzi, and D.~J. Fleet, ``Video diffusion models,'' \emph{arXiv:2204.03458}, 2022.

\bibitem{yang2023diffusion}
R.~Yang, P.~Srivastava, and S.~Mandt, ``Diffusion probabilistic modeling for video generation,'' \emph{Entropy}, vol.~25, no.~10, p. 1469, 2023.

\bibitem{blattmann2023videoldm}
A.~Blattmann, R.~Rombach, H.~Ling, T.~Dockhorn, S.~W. Kim, S.~Fidler, and K.~Kreis, ``Align your latents: High-resolution video synthesis with latent diffusion models,'' in \emph{IEEE Conference on Computer Vision and Pattern Recognition ({CVPR})}, 2023.

\bibitem{gu2022stochastic}
T.~Gu, G.~Chen, J.~Li, C.~Lin, Y.~Rao, J.~Zhou, and J.~Lu, ``Stochastic trajectory prediction via motion indeterminacy diffusion,'' in \emph{Proceedings of the IEEE/CVF Conference on Computer Vision and Pattern Recognition}, 2022, pp. 17\,092--17\,101.

\bibitem{choi2023dice}
Y.~Choi, R.~C. Mercurius, S.~M.~A. Shabestary, and A.~Rasouli, ``Dice: Diverse diffusion model with scoring for trajectory prediction,'' \emph{arXiv preprint arXiv:2310.14570}, 2023.

\bibitem{chen2023equidiff}
K.~Chen, X.~Chen, Z.~Yu, M.~Zhu, and H.~Yang, ``Equidiff: A conditional equivariant diffusion model for trajectory prediction,'' in \emph{2023 IEEE 26th International Conference on Intelligent Transportation Systems (ITSC)}, 2023, pp. 746--751.

\bibitem{rempe2023trace}
D.~Rempe, Z.~Luo, X.~B. Peng, Y.~Yuan, K.~Kitani, K.~Kreis, S.~Fidler, and O.~Litany, ``Trace and pace: Controllable pedestrian animation via guided trajectory diffusion,'' in \emph{Proceedings of the IEEE/CVF Conference on Computer Vision and Pattern Recognition}, 2023, pp. 13\,756--13\,766.

\bibitem{li2023bcdiff}
R.~Li, C.~Li, D.~Ren, G.~Chen, Y.~Yuan, and G.~Wang, ``Bcdiff: Bidirectional consistent diffusion for instantaneous trajectory prediction,'' in \emph{Advances in Neural Information Processing Systems}, A.~Oh, T.~Naumann, A.~Globerson, K.~Saenko, M.~Hardt, and S.~Levine, Eds., vol.~36.\hskip 1em plus 0.5em minus 0.4em\relax Curran Associates, Inc., 2023, pp. 14\,400--14\,413.

\bibitem{Mao2023leapfrog}
W.~Mao, C.~Xu, Q.~Zhu, S.~Chen, and Y.~Wang, ``Leapfrog diffusion model for stochastic trajectory prediction,'' in \emph{Proceedings of the IEEE/CVF Conference on Computer Vision and Pattern Recognition (CVPR)}, June 2023, pp. 5517--5526.

\bibitem{westny2024diffusion}
T.~Westny, B.~Olofsson, and E.~Frisk, ``Diffusion-based environment-aware trajectory prediction,'' \emph{arXiv preprint arXiv:2403.11643}, 2024.

\bibitem{zhong2023guided}
Z.~Zhong, D.~Rempe, D.~Xu, Y.~Chen, S.~Veer, T.~Che, B.~Ray, and M.~Pavone, ``Guided conditional diffusion for controllable traffic simulation,'' in \emph{2023 IEEE International Conference on Robotics and Automation (ICRA)}.\hskip 1em plus 0.5em minus 0.4em\relax IEEE, 2023, pp. 3560--3566.

\bibitem{guo2023scenedm}
Z.~Guo, X.~Gao, J.~Zhou, X.~Cai, and B.~Shi, ``Scenedm: Scene-level multi-agent trajectory generation with consistent diffusion models,'' \emph{arXiv preprint arXiv:2311.15736}, 2023.

\bibitem{yang2023tsdit}
C.~Yang and T.~Shi, ``Tsdit: Traffic scene diffusion models with transformers,'' \emph{arXiv preprint arXiv:2405.02289}, 2023.

\bibitem{wang2024dragtraffic}
S.~Wang, G.~Sun, F.~Ma, T.~Hu, Y.~Song, L.~Zhu, and M.~Liu, ``Dragtraffic: A non-expert interactive and point-based controllable traffic scene generation framework,'' \emph{arXiv preprint arXiv:2404.12624}, 2024.

\bibitem{zhong2023language}
Z.~Zhong, D.~Rempe, Y.~Chen, B.~Ivanovic, Y.~Cao, D.~Xu, M.~Pavone, and B.~Ray, ``Language-guided traffic simulation via scene-level diffusion,'' in \emph{Conference on Robot Learning}.\hskip 1em plus 0.5em minus 0.4em\relax PMLR, 2023, pp. 144--177.

\bibitem{chang2023editing}
W.-J. Chang, C.~Tang, C.~Li, Y.~Hu, M.~Tomizuka, and W.~Zhan, ``Editing driver character: Socially-controllable behavior generation for interactive traffic simulation,'' \emph{IEEE Robotics and Automation Letters}, vol.~8, no.~9, pp. 5432--5439, 2023.

\bibitem{pronovost2023scenariodiff}
E.~Pronovost, M.~R. Ganesina, N.~Hendy, Z.~Wang, A.~Morales, K.~Wang, and N.~Roy, ``Scenario diffusion: Controllable driving scenario generation with diffusion,'' in \emph{Advances in Neural Information Processing Systems}, A.~Oh, T.~Naumann, A.~Globerson, K.~Saenko, M.~Hardt, and S.~Levine, Eds., vol.~36.\hskip 1em plus 0.5em minus 0.4em\relax Curran Associates, Inc., 2023, pp. 68\,873--68\,894.

\bibitem{chitta2024sledge}
K.~Chitta, D.~Dauner, and A.~Geiger, ``Sledge: Synthesizing simulation environments for driving agents with generative models,'' \emph{arXiv preprint arXiv:2403.17933}, 2024.

\bibitem{janner2022planning}
M.~Janner, Y.~Du, J.~B. Tenenbaum, and S.~Levine, ``Planning with diffusion for flexible behavior synthesis,'' \emph{arXiv preprint arXiv:2205.09991}, 2022.

\bibitem{wang2022diffusion}
Z.~Wang, J.~J. Hunt, and M.~Zhou, ``Diffusion policies as an expressive policy class for offline reinforcement learning,'' \emph{arXiv preprint arXiv:2208.06193}, 2022.

\bibitem{ajay2022conditional}
A.~Ajay, Y.~Du, A.~Gupta, J.~Tenenbaum, T.~Jaakkola, and P.~Agrawal, ``Is conditional generative modeling all you need for decision-making?'' \emph{arXiv preprint arXiv:2211.15657}, 2022.

\bibitem{hansen2023idql}
P.~Hansen-Estruch, I.~Kostrikov, M.~Janner, J.~G. Kuba, and S.~Levine, ``Idql: Implicit q-learning as an actor-critic method with diffusion policies,'' \emph{arXiv preprint arXiv:2304.10573}, 2023.

\bibitem{he2023diffusion}
H.~He, C.~Bai, K.~Xu, Z.~Yang, W.~Zhang, D.~Wang, B.~Zhao, and X.~Li, ``Diffusion model is an effective planner and data synthesizer for multi-task reinforcement learning,'' \emph{arXiv preprint arXiv:2305.18459}, 2023.

\bibitem{pearce2023imitating}
T.~Pearce, T.~Rashid, A.~Kanervisto, D.~Bignell, M.~Sun, R.~Georgescu, S.~V. Macua, S.~Z. Tan, I.~Momennejad, K.~Hofmann, \emph{et~al.}, ``Imitating human behaviour with diffusion models,'' \emph{arXiv preprint arXiv:2301.10677}, 2023.

\bibitem{yang2024diffusion}
B.~Yang, H.~Su, N.~Gkanatsios, T.-W. Ke, A.~Jain, J.~Schneider, and K.~Fragkiadaki, ``Diffusion-es: Gradient-free planning with diffusion for autonomous driving and zero-shot instruction following,'' \emph{arXiv preprint arXiv:2402.06559}, 2024.

\bibitem{xiao2023safediffuser}
W.~Xiao, T.-H. Wang, C.~Gan, and D.~Rus, ``Safediffuser: Safe planning with diffusion probabilistic models,'' \emph{arXiv preprint arXiv:2306.00148}, 2023.

\bibitem{feng2024ltldog}
Z.~Feng, H.~Luan, P.~Goyal, and H.~Soh, ``Ltldog: Satisfying temporally-extended symbolic constraints for safe diffusion-based planning,'' \emph{IEEE Robotics and Automation Letters}, vol.~9, no.~10, pp. 8571--8578, 2024.

\bibitem{kondo2024cgd}
K.~Kondo, A.~Tagliabue, X.~Cai, C.~Tewari, O.~Garcia, M.~Espitia-Alvarez, and J.~P. How, ``Cgd: Constraint-guided diffusion policies for uav trajectory planning,'' \emph{arXiv preprint arXiv:2405.01758}, 2024.

\bibitem{Dauner2023CORL}
D.~Dauner, M.~Hallgarten, A.~Geiger, and K.~Chitta, ``Parting with misconceptions about learning-based vehicle motion planning,'' in \emph{Conference on Robot Learning (CoRL)}, 2023.

\bibitem{ascher1998computer}
\BIBentryALTinterwordspacing
U.~M. Ascher and L.~R. Petzold, \emph{Computer Methods for Ordinary Differential Equations and Differential-Algebraic Equations}.\hskip 1em plus 0.5em minus 0.4em\relax Philadelphia, PA: Society for Industrial and Applied Mathematics, 1998. [Online]. Available: \url{https://epubs.siam.org/doi/abs/10.1137/1.9781611971392}
\BIBentrySTDinterwordspacing

\bibitem{treiber2000congested}
M.~Treiber, A.~Hennecke, and D.~Helbing, ``Congested traffic states in empirical observations and microscopic simulations,'' \emph{Physical review E}, vol.~62, no.~2, p. 1805, 2000.

\bibitem{nuplan}
K.~T. e.~a. H.~Caesar, J.~Kabzan, ``Nuplan: A closed-loop ml-based planning benchmark for autonomous vehicles,'' in \emph{CVPR ADP3 workshop}, 2021.

\bibitem{trindade2007financial}
\BIBentryALTinterwordspacing
A.~A. Trindade, S.~Uryasev, A.~Shapiro, and G.~Zrazhevsky, ``Financial prediction with constrained tail risk,'' \emph{Journal of Banking \& Finance}, vol.~31, no.~11, pp. 3524--3538, 2007, risk Management and Quantitative Approaches in Finance. [Online]. Available: \url{https://www.sciencedirect.com/science/article/pii/S0378426607001392}
\BIBentrySTDinterwordspacing

\bibitem{hong2014montecarlo}
\BIBentryALTinterwordspacing
L.~J. Hong, Z.~Hu, and G.~Liu, ``Monte carlo methods for value-at-risk and conditional value-at-risk: A review,'' \emph{ACM Trans. Model. Comput. Simul.}, vol.~24, no.~4, nov 2014. [Online]. Available: \url{https://doi.org/10.1145/2661631}
\BIBentrySTDinterwordspacing

\end{thebibliography}

\addtolength{\textheight}{-12cm}   

\end{document}